%% file: main.tex
\documentclass[journal]{IEEEtran}
\ifCLASSINFOpdf
\else
\fi

\usepackage{multirow}
\usepackage{colortbl} 
\usepackage{dsfont}
\usepackage{arydshln}
\usepackage{tabularx}
\usepackage{subcaption}
\usepackage{graphicx}
\usepackage{amssymb}
\usepackage{amsmath}
\usepackage{pifont}
\usepackage{xspace}
\usepackage{bm}
\usepackage{enumitem}
\usepackage{booktabs}
\usepackage{xcolor}
\usepackage{verbatim}
\usepackage{hyperref}

\begin{document}
%
\title{Simultaneous Detection and Interaction Reasoning for Object-Centric Action Recognition}
%
%
%


\author{
	\IEEEauthorblockN{
		Xunsong Li\IEEEauthorrefmark{1}, 
		Pengzhan Sun\IEEEauthorrefmark{2}, 
		Yangcen Liu\IEEEauthorrefmark{1}, 
		Lixin Duan\IEEEauthorrefmark{1} 
		Wen Li\IEEEauthorrefmark{1}}
  
        \IEEEauthorblockA{\IEEEauthorrefmark{1}University of Electronic Science and Technology of China}
        \IEEEauthorblockA{\IEEEauthorrefmark{2}National University of Singapore}

        {\tt\small \{lixunsonghcl, liuyangcen112358, lxduan, liwenbnu\}@gmail.com, pengzhan@comp.nus.edu.sg}
}

\maketitle

\newcommand{\modelname}{DAIR\xspace}
\newcommand{\moduleA}{PatchDec\xspace}
\newcommand{\moduleB}{IRA\xspace}
\newcommand{\ie}{\emph{i.e., }}
\newcommand{\eg}{\emph{e.g., }}
\newcommand{\etal}{\emph{et al.}}
\newcommand{\etc}{\emph{etc.}}
\newcommand{\wrt}{\emph{w.r.t. }}
\newcommand{\cf}{\emph{cf. }}
\newcommand{\vs}{\emph{v.s. }}
\newcommand{\aka}{\emph{aka. }}
\newcommand{\secref}[1]{Section \ref{#1}}
\newcommand{\w}[0]{\texttt{w}}
\newcommand{\cmark}{\ding{51}}
\newcommand{\xmark}{\ding{55}}
\newcommand{\tablestyle}[2]{\setlength{\tabcolsep}{#1}\renewcommand{\arraystretch}{#2}\centering\footnotesize}

\newcommand{\xs}[1]{\textcolor{blue}{#1}}
\newcommand{\pz}[1]{\textcolor{red}{#1}}
\newcommand{\yc}[1]{\textcolor{orange}{#1}}

\begin{abstract}

The interactions between human and objects are important for recognizing object-centric actions. Existing methods usually adopt a two-stage pipeline, where object proposals are first detected using a pretrained detector, and then are fed to an action recognition model for extracting video features and learning the object relations for action recognition.
However, since the action prior is unknown in the object detection stage, important objects could be easily overlooked, leading to inferior action recognition performance. 
In this paper, we propose an end-to-end object-centric action recognition framework that simultaneously performs \underline{D}etection \underline{A}nd \underline{I}nteraction \underline{R}easoning (dubbed \modelname) in one stage. Particularly, after extracting video features using a base network, we design three consecutive modules for simultaneously learning object detection and interaction reasoning. Firstly, we build a Patch-based Object Decoder (\moduleA) to generate object proposals from video patch tokens. Then, we design an Interactive Object Refining and Aggregation (\moduleB) to identify the interactive objects that are important for action recognition. The \moduleB module adjusts the interactiveness scores of proposals based on their relative position and appearance, and aggregates the object-level information into global video representation. Finally, we build an Object Relation Modeling (ORM) module to encode the object relations. These three modules together with the video feature extractor can be trained jointly in an end-to-end fashion, thus avoiding the heavy reliance on an off-the-shelf object detector, and reducing the multi-stage training burden.
We conduct experiments on two datasets, Something-Else and Ikea-Assembly, to evaluate the performance of our proposed approach on conventional, compositional, and few-shot action recognition tasks. Through in-depth experimental analysis, we show the crucial role of \textit{interactive} objects in learning for action recognition, and we can outperform state-of-the-art methods on both datasets.
We hope our \modelname can provide a new perspective for object-centric action recognition. The code is available at
\href{https://github.com/lixunsong/DAIR}{https://github.com/lixunsong/DAIR}.
\end{abstract}

%
\IEEEpeerreviewmaketitle

\input{sections/1_introduction}
\input{sections/2_related_work}
\input{sections/3_method}
\input{sections/4_experiment}
\input{sections/5_conclusion}


\ifCLASSOPTIONcaptionsoff
  \newpage
\fi



%

\bibliographystyle{IEEEtran}
\bibliography{refer}

\end{document}

%% file: sections/1_introduction.tex
\section{Introduction}

\begin{figure}
    \centering
    \includegraphics[width=0.95\linewidth]{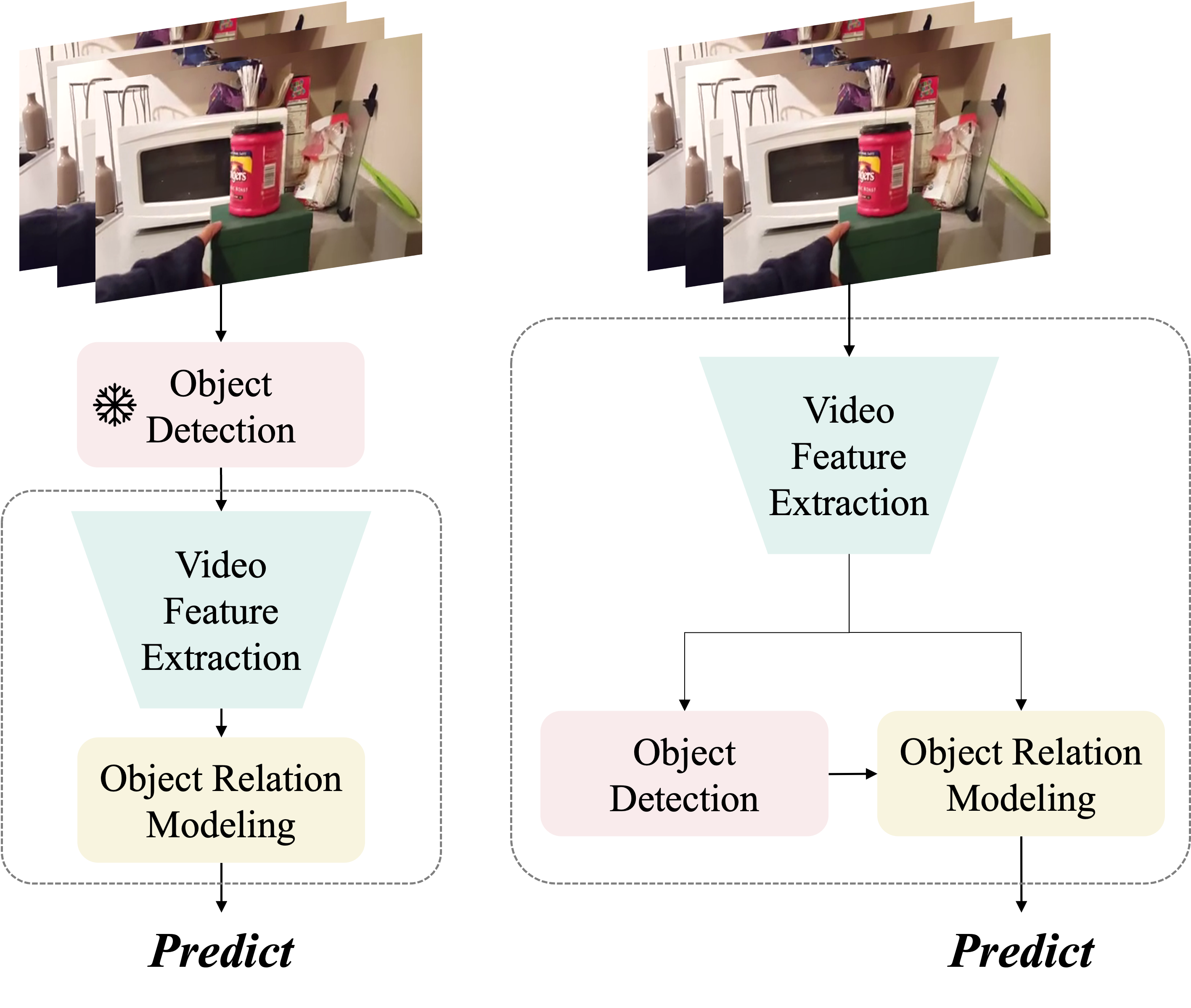}
    \caption{Comparison between the previous two-stage pipeline (left figure) and our proposed \modelname (right figure). Previous works limited their research scope to only video feature extraction and object relation modeling, which relied on the output of an off-the-shelf object detector in the first stage. Our proposed \modelname incorporates object detection into the end-to-end model architecture, not only reducing multi-stage training cost but also allowing the object detector to focus on interactive objects in the action.}
    \label{fig:intro}
\end{figure}
\vspace{-4pt}

Significant progress has been made in action recognition tasks~\cite{li2018unified} with the advancement of video representation learning, such as video convolution models~\cite{carreira2017quo, feichtenhofer2019slowfast, simonyan2014two, lin2019tsm, wang2016temporal} and video transformers~\cite{li2022mvitv2, arnab2021vivit, patrick2021keeping}. However, these models mainly extract spatio-temporal representations from the scene level, which can result in spurious correlations~\cite{choi2019can} learned from background information. Recent research~\cite{herzig2022object, sun2021counterfactual, kim2021motion, yan2023progressive, materzynska2020something, ben2022bringing} focuses on a more efficient method: understanding video actions from human-object interaction information, which is still an under-explored topic.

Learning the interaction~\cite{xu2019interact} between human and objects is a crucial factor in understanding video actions. 
Actions often share inherent physical properties of objects, such as their relative position movements, appearance changes, and interaction with environments. Therefore, building an object-centric action recognition system could lead to more efficient action learning, and also allows for better generalization to novel objects and scenes~\cite{materzynska2020something, zhang2022object}.

To incorporate object-level information into video representations, most existing works~\cite{tmmhoi, tmmhuman, herzig2022object, sun2021counterfactual, kim2021motion, yan2023progressive, materzynska2020something, ben2022bringing} adopt a separate architecture: 1) obtaining object proposals and features from offline detectors or manual labeling, 2) performing spatial-temporal relationship modeling based on these object proposals.
Most existing methods primarily focus on the latter phase of object relationship modeling and employ various ways for better interaction modeling in action recognition, such as multi-branch architectures~\cite{kim2021motion, sun2021counterfactual, radevski2021revisiting}, graph networks~\cite{wang2018videos, tmmpose,ou2022object} and transformer models~\cite{herzig2022object, ben2022bringing}.
However, adopting the two-stage recognition pipeline still faces some critical issues in practical applications. Firstly, the requirement of multi-stage training for the object detector and the action classification entails a large computation cost. Secondly, given that these two models are trained independently, the performance of the downstream action recognition model heavily relies on the quality of object detections. This reliance is clearly observed in state-of-the-art approaches~\cite{herzig2022object, materzynska2020something, radevski2021revisiting}, where a significant performance gap exists between using oracle object annotations and using detected results.

In this work, we build an end-to-end framework that simultaneously Detects Active objects and conducts Interaction Reasoning (namely \modelname) for action recognition.
\modelname eliminates the reliance on additional object detectors and enhances the capability of interaction modeling by better exploring active objects.
Our novel framework is composed of three parts: First, the PatchDec module (Patch Decoder) extracts video features and instance proposals. Second, the IRA module (Interactive Object Refining and Aggregation) performs instance relation reasoning. Finally, the ORM module (Object Relation Modeling) aggregates the patch-level video feature and instance-level relation feature.
We compare the traditional two-stage pipelines and ours in Figure~\ref{fig:intro}. It is worth noting that the overall framework is versatile and can integrate with various existing video backbones and object relation modeling methods.

\begin{figure}[tb]
    \centering
    \includegraphics[width=0.95\linewidth]{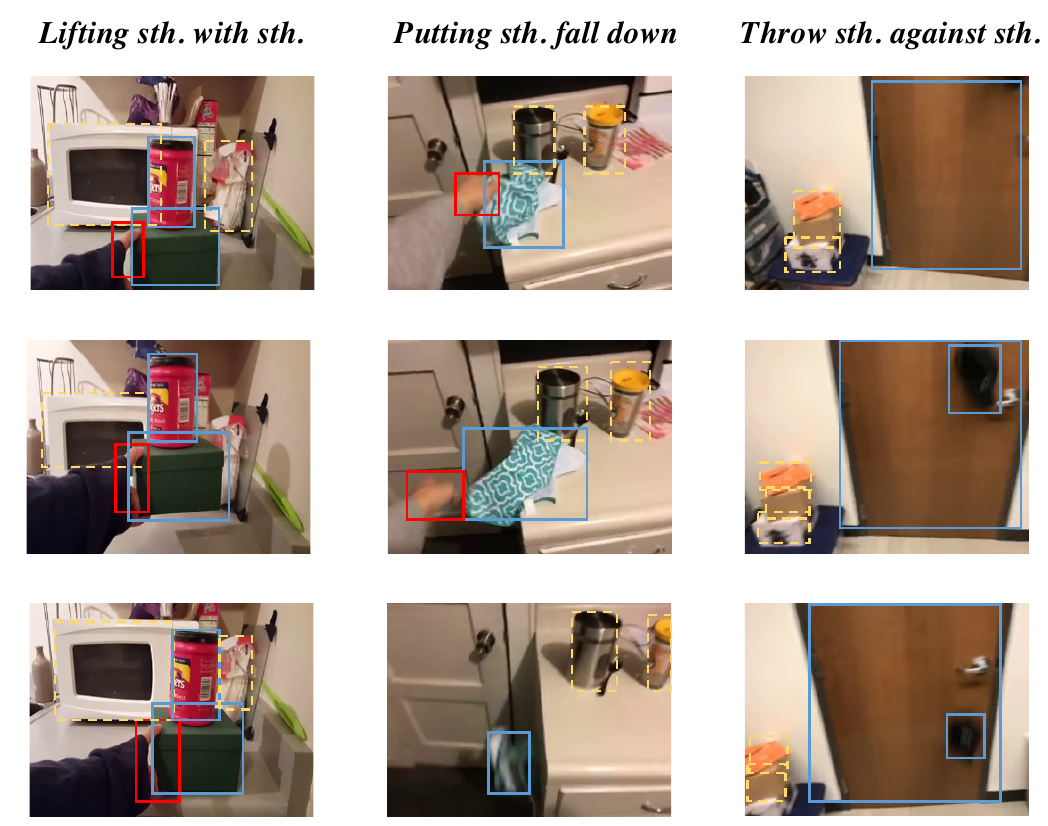}
    \caption{Illustration of interactive (\textcolor[HTML]{3399FF}{blue} boxes), non-interactive objects (\textcolor[HTML]{FFD333}{yellow} dashed boxes) and subjects (\textcolor{red}{red} boxes) in the action. The non-interactive objects could be the blur to action classification.}
    \label{fig:intro2}
    
\end{figure}

Compared with the detection in static images, detecting objects in video actions further requires finding the ``interactive'' objects that are relevant to the action.
Figure~\ref{fig:intro2} shows some clips that involve multiple objects in the scene. The yellow boxes correspond to background objects. Although they may appear as salient regions and can be easily detected by image-based detectors, they do not contribute to understanding the action and can even introduce additional noise. Therefore, how to accurately focus on only the interactive objects is an important issue for understanding human-object actions in the video. 

To this end, we enhance video object detection by incorporating interactiveness knowledge at patch-level and instance-level. At the patch-level, we follow DETR~\cite{carion2020end} that utilizes a set of object queries to attend to the video patch tokens, which are then used to predict both subjects (e.g., the person or hand) and objects in the scene. Since the patch tokens integrate spatial-temporal contexts, they enhance their sensitivity towards moving object regions. Based on the detection results, we can extract instance-level features of objects and subjects, such as their appearance and relative position features.

At the instance-level, we propose \moduleB, which takes the detected subjects and object proposals aforementioned as input. \moduleB models the layout relations and appearance changes of subjects and objects. Specifically, ~\moduleB recalibrates the interactiveness scores of candidate objects, particularly to determine the involvement of \textit{passive} objects in the action. Using the interaction scores predicted by the IRA module as weights, we consolidate object-centric information from each frame. We then integrate this information into global video representations with our custom ORM. We accomplish this by concatenating these aggregated features with video patch tokens.

Coupling these designs with a video backbone, we are able to establish an end-to-end object-centric action recognition model that significantly improves the video backbone by introducing only a few parameters and computational overhead. Specifically, our approach is evaluated on two challenging datasets: Something-Else~\cite{materzynska2020something} and IKEA-Assembly~\cite{ben2021ikea}, and has achieved a $3.5$\% top-1 accuracy improvement on Something-Else and a $5.6$\% macro-recall promotion on Ikea-Assembly compared with the baseline model MViTv2~\cite{li2022mvitv2}. Moreover, our model outperforms state-of-art object-centric methods~\cite{herzig2022object, ben2022bringing, kim2021motion} on both datasets, without the requirement of additional detectors or external data.

Our contributions can be summarized as follows:

\begin{itemize}[topsep=0pt, itemsep=0pt]
    \item We present \modelname, an end-to-end framework that can detect active objects and conduct interaction reasoning simultaneously. This novel framework employs a shared video encoder for both tasks, eliminating the requirement for external detectors and enabling integration with various video transformers in a versatile manner.
    
    \item We demonstrate the importance of discerning interactive objects in object-centric action learning, and devise the PatchDec and IRA modules that detects interactive objects at both patch-level and instance-level. This enhances the conventional video representation with more accurate interactive object information, thereby improving the generalization ability of the action recognition model to unseen scenes and objects.
    
    \item We evaluate our framework on conventional, compositional, and few-shot action recognition tasks across two challenging datasets, Something-Else~\cite{materzynska2020something} and Ikea-Assembly~\cite{ben2021ikea}. The experimental results show that our approach outperforms other existing works, reaching a new state-of-the-art. 
    
\end{itemize}
 
\vspace{-4pt}

%% file: sections/2_related_work.tex
\section{Related Work}

\noindent\textbf{Object-centric Video Models.} Integrating object-level information into conventional video models (\eg convolutional networks~\cite{carreira2017quo, feichtenhofer2019slowfast, feichtenhofer2020x3d} and transformers~\cite{li2022mvitv2, patrick2021keeping, arnab2021vivit}) has been extensively explored. Most studies follow a separate architecture that obtains object bounding boxes from an auxiliary detector or manual annotations and then explores various reasoning modules or fusion strategies for capturing the object-level interaction in videos.
Object Relation Network~\cite{baradel2018object} performs object relation reasoning on two consecutive frames with GRU~\cite{cho2014learning} units. STRG~\cite{wang2018videos} builds a spatio-temporal graph over object features that are extracted from a region proposal network (RPN)~\cite{he2017mask} over I3D features. OR2G~\cite{ou2022object} further considers the pairwise relationship as graph nodes and captures higher-order relation transformation for action recognition. 
In addition to leveraging ROI features for object-centric representation, STIN~\cite{materzynska2020something} takes object bounding boxes as input and models the relative position changes of objects. This approach demonstrates robust action recognition performance even when operating on unseen objects. Meanwhile, STLT~\cite{radevski2021revisiting} uses a transformer encoder to enhance the bounding box based action representation. To effectively combine multiple types of object information, such as appearance and positions, CDN~\cite{sun2021counterfactual} uses a multi-branch network to concatenate the features from each branch before the classification head. PIFL~\cite{yan2023progressive} progressively fuses object-centric information and performs interaction reasoning simultaneously. ORViT~\cite{herzig2022object} introduces a transformer block that injects object appearance and motion information into video patch tokens through joint self-attention, and fuses them from the shallow layers of the transformer model for better effects.

Apart from the utilization of the object-centric information side, ~\cite{ben2022bringing} addresses the expensive cost of annotating dense structure in video data and proposes to align external image data with structure annotations to video data. Their approach does not require object annotations or external detectors but requires good alignment between the video and image data (\eg the number of subjects and objects). Besides, it cannot guarantee the discovery of interactive objects. In contrast, we provide a more unified approach to bridge existing video models and interaction reasoning models for object-centric action recognition.

\begin{figure*}[tpb]
    \centering
    \includegraphics[width=1.0\linewidth]{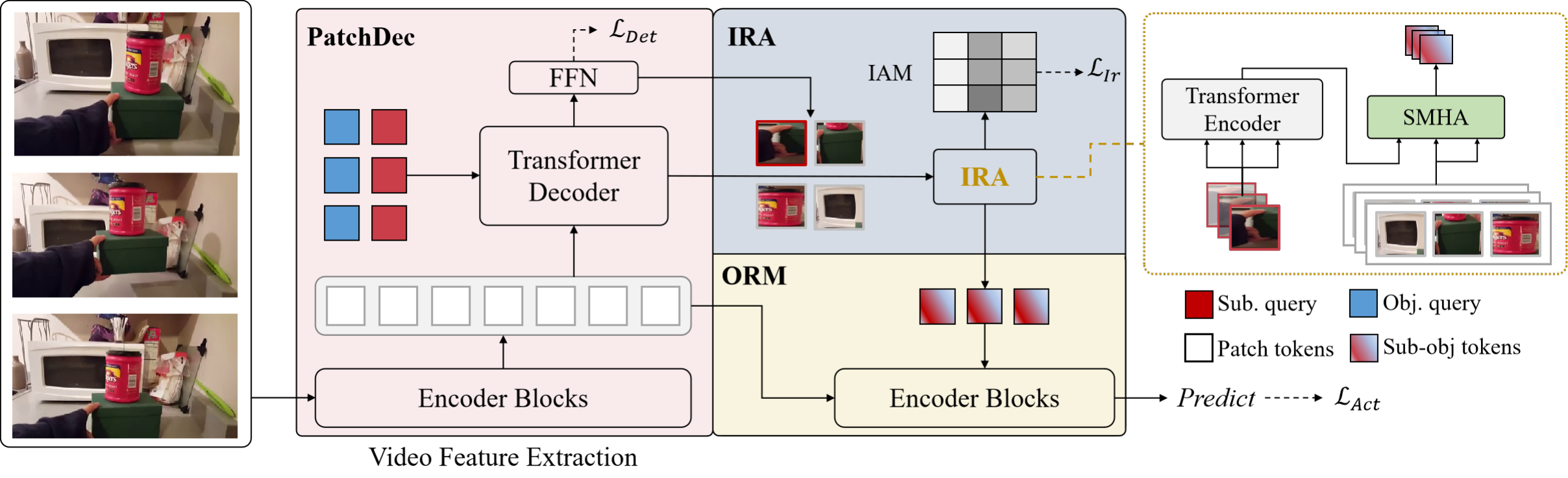}
    \caption{Overview of the proposed \modelname. It consists of three main components and takes dense sample frames as a transformer decoder implements input.~\moduleA (Patch Decoder, can be instantiated by any video transformer), which responds to decoding objects from learnable queries after attending with patch tokens. ~\moduleB (Interactive Object Refining and Aggregation), which takes the instance-level representations of subjects and object proposals as input, and performs inter-relation reasoning to refine the confidence scores of objects.  In ORM (Object Relation Modeling), subject and object features are comprised, and we take the [CLS] token as the final video representation to predict actions.}
    \label{fig:framework}
\end{figure*}
\vspace{-3pt}

\noindent\textbf{Video Object Detection.}
Extending the detection capability from static images~\cite{qiu2020hierarchical} to videos~\cite{li2020spatio} is widely explored. The conventional VOD task requires detecting objects at each frame and linking the same object across frames. Its core issue lies in how to aggregate temporal contexts to enhance frame-wise detection accuracy. A variety of approaches are proposed to facilitate the information interchanges across frames, such as warping with optic flow models~\cite{zhu2017flow, zhu2018towards, wang2018fully}, integrating with memory mechanism~\cite{tmmvod, jiang2019video,tmmvideoobjectdetection, deng2019object, chen2020memory} or recurrent networks~\cite{chen2020memory, deng2019object, guo2019progressive}. However, VOD focuses on how to extract all foreground regions (\ie objects) that appear in the scene, rather than on the objects of interest only related to the action.

A particular line of work has emerged in detecting interacted objects in videos, which extends Human-Object Interaction (HOI) learning from images to videos. This entails localizing the objects involved in the interaction and identifying their interaction categories within a given video clip. The video-based HOI (Vid-HOI) task can be broadly categorized into two types, based on the granularity of their predictions: (1) clip-level Vid-HOI, wherein the tracklet of each object is first obtained and then aggregated to form a compact representation for interaction recognition~\cite{tu2022video,chiou2021sthoi,li2022discovering}; and (2) frame-level Vid-HOI, wherein the interaction prediction is performed at each frame and is typically incorporated with temporal contexts using GNNs~\cite{qi2018learning} or transformers~\cite{ji2021detecting}. Our object-centric action recognition model adopts a similar process to the clip-level paradigm, but some differences remain. Firstly, we perform object detection based on the features obtained from the video model, which contains rich dense spatio-temporal information. On the other hand, HOI methods use an image encoder to process individual frames and obtain visual features, which are then used for tracking and temporal reasoning. Secondly, HOI methods typically detect a limited set of object categories and assume that each active object corresponds to a certain predicate. However, general video action recognition typically involves multiple objects, requiring subsequent object relationship modeling. This is outside the scope of HOI models.

%% file: sections/3_method.tex
\section{Method}
\textbf{Overview. } As shown in Figure~\ref{fig:framework}, we present a novel object-centric action recognition pipeline for Detecting Active objects and conducting Interaction Reasoning (\modelname), which is able to train and infer in an end-to-end manner.
Our model takes in densely sampled video clips with frame-wise object annotation provided during the training phase. Remarkably, our model does not require additional object annotations during the testing phase. 
\modelname~streamlines the recognition pipeline with the components of video feature extraction, patch-based object decoding, interactive object refining, and object relation modeling. 
\textit{First}, the input video is encoded by a backbone model to 3-D features as detailed in Section~\ref{sec:3.1}.
\textit{Second}, in PatchDec, the spatial-temporal feature is patchified into a series of patch tokens, which are then updated through several self-attention blocks.
We initialize some object queries and send them into the transformer decoder along with video patch tokens. These updated queries are responded to predict both subjects and objects from the whole scene as detailed in Section~\ref{sec:3.2}. 
Then, we refine the interactiveness score of detected candidate objects with the IRA module, by considering their spatio-temporal relationship with subjects as depicted in Section~\ref{sec:3.3}. Object-centric information is aggregated at each frame according to the interactiveness score.
\textit{Finally}, we concatenate the aggregated object information with the global context of videos and propagate them through self-attention layers. In this way, the object relation are modeled and injected into the video representations. We take the final [CLS] token for action classification.

\subsection{Video Feature Extraction}\label{sec:3.1}

Our approach utilizes a video transformer for encoding input videos and extracting spatio-temporal representations. 
Specifically, given a video clip $\bm{V} \in R^{T \times 3 \times H\times W}$, we utilize a shallow 3D convolution layer to generate down-sampled feature maps, which are then unfolded along both spatial and temporal axes to produce patch tokens. Then, each token is added with spatio-temporal position embeddings $\bm{E}_{pos}$: 
\begin{equation}
    \bm{Z} = \text{PatchEmb}(\bm{V}) + \bm{E}_{\text{pos}}.
\end{equation}
Here $Z$ denotes video patch tokens. Then, these tokens are fed into several stacked self-attention blocks to repeatedly update their representation. We denote the output of the $i$-th layer as $F_{i}\in R^{L_i\times D_{i}}$, where $L_i$ denotes the sequence length and $D_i$ is the channel number. This process can be formulated as:

\begin{equation}
    \bm{F}_{(i)} = \begin{cases}
        \bm{Z}, & i = 0 \\
        \text{Enc}_{(i)}(\bm{F}_{(i-1)}), & i > 0
    \end{cases}
\end{equation}

It is worth noting that our approach is agnostic to the specific choice of video transformer model. In our experiments, we choose to use MViTV2~\cite{li2022mvitv2} as the video encoding backbone due to its good empirical performance.

\subsection{Patch-based Object Decoding}\label{sec:3.2}

In PatchDec, we adopt query-based transformer decoders, following the approach of transformer-based detectors such as~\cite{zhu2020deformable, carion2020end, tamura2021qpic}, to localize and identify objects in videos. By doing so, we randomly initialize a set of learnable queries that aggregate information from the visual feature through cross-attention operations. The output feature of each object query from the decoder is then transformed by a Feed-Forward Network (FFN) to predict the object's location and classification score. Unlike image-based transformer detectors, the transformer encoder and decoder of our approach capture temporal context information, making it more focused on local motion patterns. This allows our object decoder to distinguish between moving and stationary objects and increases the probability of identifying interactive objects. 

\textbf{Query Initialization.}\label{sec:3.1.2}
For each frame, we initialize $N_{q}$ queries and each query consists of three types of embeddings: identity embedding, spatial position embedding, and temporal embedding. The detail and intuition of each type of embedding are described as follows.

\textbf{Identity Embedding.} The identity embedding indicates whether the query predicts the subject or object. This is done for two reasons: firstly, distinguishing between the subject and object of interaction serves as a vital cue for identifying actions; secondly, since the visual variability of objects is typically larger than that of subjects, processing them separately can enhance the accuracy of subject predictions. Typically, half of the queries at each frame are assigned with subject embeddings while the remainder with object embeddings.

\textbf{Spatial Embedding.} Following prior work on transformer detectors~\cite{wang2022anchor, zhu2020deformable}, we assign each query a reference point and predict its relative offsets. This approach has been shown to expedite the training convergence of detectors~\cite{wang2022anchor, zhu2020deformable}.
To avoid the predictions of queries at different frames collapsing to the results of a certain single frame, we independently sample the coordinates for each frame with a uniform distribution from 0 to 1. We project these 2-D coordinates into d-dimensional vectors through an MLP to align the dimensions. 

\textbf{Temporal Embedding.} The temporal order is important for action understanding. We add each frame's queries with the temporal position embeddings to distinguish the same object present in different frames. We reuse the temporal position embedding previously used for patch tokens, allowing the queries in each frame to attend to the visual features of their corresponding frames.

\textbf{Object Decoding.}\label{sec:3.1.3} We take the output of the $l$-th layer of the video transformer as visual features and feed them with the aforementioned queries into $N$ transformer decoder layers. These queries are updated after the self-attention with themselves and cross-attention with visual features. Notably, the attention computation is applied across the space-time dimensions. We employ four independent predictors composed of FFNs to generate the coordinates and category predictions for both subjects and objects: 
\begin{align}
\bm{\hat{b}}^s &=\text{FFN}_{sbox}(\operatorname{Dec}(\bm{q}_s,\bm{F}_l)), \\
\bm{\hat{b}}^o &= \text{FFN}_{obox}(\operatorname{Dec}(\bm{q}_o, \bm{F}_l)), \\
\hat{c}^s &= \sigma(\text{FFN}_{sc}(\operatorname{Dec}(\bm{q}_s, \bm{F}_l))), \\
\hat{c}^o &= \sigma(\text{FFN}_{oc}(\operatorname{Dec}(\bm{q}_o, \bm{F}_l))),
\end{align}
where $\bm{q}_s$ and $\bm{q}_o$ represent the queries of subjects and objects respectively, $\sigma$ is the sigmoid function, and $\text{FFN}_{sbox}$, $\text{FFN}_{obox}$, $\text{FFN}_{sc}$, $\text{FFN}_{oc}$ response for decoding subject-box-offsets $\bm{\hat{b}}^s$, object-box-offsets $\bm{\hat{b}}^o$, subject-confidence-score $\hat{c}^{s}$ and object-class $\hat{c}^o$ respectively. The semantic difference between the learned Sub. and Obj. query can be guided in two folds. First, our dataset provides annotations for subjects (hands or humans) and objects, allowing us to distinguish between them during training. Second, the Sub. and Obj. queries are initialized independently and optimized separately with the set prediction. The two factors help to maintain the semantic distinction between the Sub. Query and Obj. Query throughout the learning process. Here we omit the subscripts $t$ used to index different frames. Following~\cite{zhu2020deformable}, the final predicted box for each query is obtained by adding the offset $\bm{\hat{b}}^{\{s, o\}}$ with its initial reference point $(q_x, q_y)$, \ie $\bm{b}^{\{s, o\}} = \{\sigma(\hat{b}_x^{\{s, o\}} + \sigma^{-1}(q_x)),  \sigma(\hat{b}_y^{\{s, o\}} + \sigma^{-1}(q_y)), \sigma(\hat{b}_w), \sigma(\hat{b}_h) \}$.

\subsection{Interactive Object Refining and Aggregation}\label{sec:3.3}
As aforementioned, queries that attend to spatio-temporal patch tokens are more sensitive to moving objects, which are \textit{salient} in the entire clip. However, in many cases, actions involve multiple objects with varying degrees of interaction, including objects that may not have direct contact with the subject or exhibit significant motion.
Take the third column in Figure~\ref{fig:intro2} as the example, in which the action label is \textit{Throw sth. against sth.}. We can observe that the ``door'' and the ``box'' next to it are both in the background, which are hard to determine whether they are involved in the action individually. However, by considering the relative positional relationship with the thrown object, it can be inferred that the ``door'' is more likely to be involved in the action than the ``box'', since the thrown object is moving toward it. 
In addition to utilizing the positional relationships between objects to infer their interactiveness, the appearance differences among instances also aid in this determination (\eg an open box is an active object that appears distinct from a closed one.).

Based on the analysis above, we introduce the~\moduleB (Interactive Object Refining and Aggregation) module to calibrate the interactiveness scores of object proposals and aggregate their visual features. It takes in both subject and object's features and employs an encoder-decoder architecture to reason their inner-relationship for interactiveness prediction. 

\textbf{Feature Construction.}\label{sec:3.2.1} Both the representations of subjects and objects are constructed based on their appearance features and spatial positions. We first rank the predicted boxes with their confidence scores in the last stage and select a fixed number of subject and object boxes. With these boxes and video patch tokens $\bm{F}_{l}$, we adopt RoIAlign~\cite{he2017mask} operation followed by an MLP to obtain object appearance features:
\begin{equation}
    \bm{v}_A^{\{s, o\}} = \operatorname{MLP}(\operatorname{RoIAlign}(\bm{\hat{b}}^{\{s, o\}}, \bm{F}_{l})).
\end{equation}
We also obtain the spatial position feature by applying an MLP on the box coordinates, resulting in $v_P^{\{s, o\}}$:
\begin{equation}
    \bm{v}_P^{\{s, o\}} = \operatorname{MLP}(\bm{\hat{b}}^{\{s, o\}}).
\end{equation}
We sum up the appearance and spatial position feature of each subject/object as its final merged representation and refer it as the \textit{sub-obj token} $\bm{v}_M^{\{s,o\}}$.

\textbf{Module Implementation.}\label{sec:3.2.2} The architecture of~\moduleB is shown in the top right part of Figure~\ref{fig:framework}, which consists of a transformer encoder and a sigmoid multi-head attention layer (SMHA). The transformer encoder processes subject tokens solely across frames to capture their tracklet movements and appearance changes. The updated subject tokens are sent into a decoder layer together with object tokens. In detail, it takes object tokens as $K$, $V$, and the updated subject tokens as $Q$. The cross-attention is computed within each frame. 
Following~\cite{liu2022interactiveness}, the original softmax function used to generate attention scores is replaced by a sigmoid function following averaging, which is suitable for the scenarios where multiple active objects are present in an action, and all of their attention scores should be $1$. The attention map $A \in R^{T\times N_o}$ is considered as the Interactiveness Assignment Matrix (IAM) of objects, where $N_o$ is the number of candidate objects at each frame. We optimize the IAM to achieve optimal assignments, as described in Section~\ref{sec:3.2}.

\subsection{Object Relation Modeling}\label{sec:3.4}
We refer to the output of the IRA module as \textit{Sub-obj} tokens, which comprise subject tokens and weighted object features, representing the interaction's representation at each frame. We concatenate these \textit{Sub-obj} tokens with video patch tokens and feed them together into the self-attention layers:

\begin{equation}
    F_{l+1} = \text{Enc}_{(i)}(concat(F_{l}, \bm{v}_F^{\{s,o\}})),
\end{equation}
where $l$ is the selected layer and LN denotes the LayerNorm operation. $F_{l}$ denotes the feature of the plugged layer in the backbone model.
Then, the interaction representation is propagated through the remaining self-attention layers of the video transformer backbone to model the relation of all tokens, as the similar approach adopted in~\cite{herzig2022object}.

\subsection{Training Objectives}\label{sec:3.4}
The training loss is composed of three components, namely, action classification loss $\mathcal{L}_{Act}$, object detection loss $\mathcal{L}_{Det}$ for~\moduleA, and the interactiveness loss $\mathcal{L}_{Ir}$ for optimizing the assign matrix of~\moduleB. The three losses are summed together and optimized jointly as: 
\begin{equation}
    \mathcal{L} = \mathcal{L}_{Act} + \lambda_{Det} \mathcal{L}_{Det} + \lambda_{Ir} \mathcal{L}_{Ir},
\end{equation} 
where $\lambda_{Det}$, $\lambda_{Ir}$ are hyper-parameters for weighting each loss. We detail the calculation of each loss term as follows.
 
\textbf{Action Classification Loss.} We take the [CLS] token from the object relation modeling block as the entire video representation and feed it to a fully connected layer to predict logit $\hat{Y}$. The action classification loss $\mathcal{L}_{Act}$ is computed by a standard cross-entropy loss between the predicted logits $\hat{Y}$ and true labels $Y$. 
\begin{equation}
    \mathcal{L}_{Act} := \operatorname{CE}(\operatorname{SoftMax}(\hat{Y}), Y)
\end{equation}

\textbf{Object Detection Loss.} For PatchDec, we follow the query-based transformer detectors to assign a bipartite matching between outputs of the object decoder and ground-truth object annotations.
This matching and optimization process is conducted separately for subjects and objects. To be concise, we outline the computation process for objects, which is the same as that of subjects.
The matching cost is composed of the box regression loss 
($L1$ loss) $\mathcal{L}_b$, the generalized IoU (Intersection Over-Union) loss~\cite{rezatofighi2019generalized} $\mathcal{L}_u$ and the binary cross-entropy classification loss $\mathcal{L}_c$. We denote object prediction results as $\mathcal{O}=\left\{o^i\right\}_{i=1}^{N_{q}}$, and $N_q$ is the total number of object queries. The cost between $i$-th ground truth $g^i$ and $\omega (i)$-th is calculated as:
\begin{equation}
\label{eq:cost}
\mathcal{L}_{\text {cost}}(g^i, o^{\omega(i)})=\lambda_b  \mathcal{L}_b^i+\lambda_u  \mathcal{L}_u^i+ \lambda_c  \mathcal{L}_c^i,
\end{equation}
where $\omega (i)$ is the index of object predictions assigned to the $i$-th ground truth, and $\lambda_b$, $\lambda_u$, $\lambda_c$ are weighting coefficients. By leveraging the Hungarian Matching algorithm~\cite{kuhn1955hungarian}, the optimal assignment $\hat{\omega}$ is determined, which achieves the minimum cost between predictions and ground truth. We utilize the same form as Eq.~\ref{eq:cost}, but with the optimal permutation $\hat{\omega}$ to calculate $\mathcal{L}_{OD}$ for training back-propagation:
\begin{equation}
\mathcal{L}_{Det}=\sum_{i}\mathcal{L}_{\text{cost}}(g^i, o^{\hat{\omega}(i)}).
\end{equation}

\textbf{Interactiveness Loss.}
As aforementioned, each entry in the assigned matrix $A$ denotes the interactiveness score of a certain object proposal. 
Since the video datasets~\cite{materzynska2020something, ben2021ikea} only provide annotations on active objects, we can directly treat the annotated objects as interactive. This means that the ground truth targets are the same for calculating $\mathcal{L}_{Det}$ and $\mathcal{L}_{Ir}$ (\ie the same set of object boxes).
The computation of $\mathcal{L}_{Ir}$ also follows the set-prediction paradigm as for $\mathcal{L}_{OD}$, except for two differences. Firstly, since we choose top-$k$ object proposals as input, the permutation $\omega(i)$ between queries and ground truth will be recalculated. Secondly, we copy the proposal boxes for calculating the box regression and IoU loss when performing matching. After obtaining the optimal one-to-one matching between top-$k$ object proposals and ground truth, $\mathcal{L}_{Ir}$ is computed as the binary classification cross-entropy loss:
\begin{equation}
  \mathcal{L}_{Ir} = \sum_t \sum_i\left(-a_t^i \cdot \log (\hat{a}_t^{\hat{\omega}(i)}) + (1-a_t^i) \cdot \log (1-\hat{a}_t^{\hat{\omega}(i)})\right),
\end{equation}
where $a_t^i \in \{0, 1\}$ refers whether the $i$-th ground truth in $t$-th frame corresponds to an active or inactive object, and $\hat{\omega}$ is the optimal permutation of proposals.

%% file: sections/4_experiment.tex
\section{Experiment}

\subsection{Datasets and Evaluation Metrics}
Following the recent object-centric action recognition methods, we evaluate our model on the two public benchmark datasets: \textbf{Something-Else}~\cite{materzynska2020something} and \textbf{IKEA-Assembly}~\cite{ben2021ikea}.

\textit{Something-Else}~\cite{materzynska2020something} is built on the basis of SSV2~\cite{goyal2017something}. It provides new data splits based on the combination of objects and actions and establishes two tasks: ``compositional'' and ``few-shot'' action recognition, which better evaluate a model's intrinsic action modeling ability, as opposed to overfitting to object appearance information. In the compositional task, there are $174$ atomic actions and 54,919 videos for training, and 57,876 videos for testing. In the few-shot task, the original $174$ actions are divided into the base set with $88$ actions and the novel set with $86$ actions. Models are pre-trained on the base set, which has a total number of $112,397$ videos, and then fine-tuned on the novel set using either $5$ or $10$ samples per class (\ie 5-shot or 10-shot). During the fine-tune stage, all model parameters are frozen except for the last classification layer. Interacted objects are annotated with bounding boxes at each frame. Following~\cite{materzynska2020something}, we evaluate our approach on this dataset with the standard classification protocol and measure both top-1 and top-5 accuracy.

\textit{IKEA-Assembly}~\cite{ben2021ikea} is a realistic furniture assemblies video dataset with a more granular scale of action definitions. The dataset includes 12 actions and 7 objects in the dataset, with a total number of 16,764 video samples. All videos are recorded from a third-person perspective. Object information are manually annotated about $1\%$ of total frames (selected keyframes), and annotations of rest frames are automatically generated from a Mask-RCNN~\cite{he2017mask} detector, which is finetuned on those keyframes. Due to a severe class imbalance issue, we adopt the micro averaged accuracy (micro) and mean of per class recall (macro-recall) metrics, following the approach in~\cite{kim2021motion, yan2023progressive}, to evaluate the performance.

\subsection{Implementation Details}
\textbf{Network configurations.} Our proposed framework is general and can be plugged into the most recent video transformer models. In our experiments, we choose MViT-S~\cite{li2022mvitv2} as the base model due to its advanced performance and initialize its parameters from a Kinetic-400 pre-trained model~\cite{kay2017kinetics}.
The~\moduleA~and~\moduleB~are inserted after the 9-th self-attention block of MViT-S~\cite{li2022mvitv2}, which is the last layer of the third scale stage with the spatial resolution $14\times 14$. We set the number of queries $N_q=40$ at each frame. The number of decoder layers for~\moduleA is $6$, and $3$ encoder layers along with $1$ decoder layer for~\moduleB. For Something-Else, we register the number of object categories as $2$ (\ie hand, object). For Ikea-Assembly, we register 7 object categories following their original definition. Since the number of active objects varies in different datasets, we choose top-$4$ object proposals as the input of~\moduleB~in Something-Else while top-$6$ proposals in Ikea-Assembly. 

\textbf{Losses and optimizers.} We use AdamW~\cite{loshchilov2017decoupled} optimizer with weight decay $1\times 10^{-4}$ for all experiments. The loss weight $\lambda_{OD}$ and $\lambda_{IA}$ are 1 and 10 respectively. In addition, the weight $\lambda_b$, $\lambda_u$, $\lambda_c$ in Eq.~\ref{eq:cost} are 5, 2, and 5 respectively.      

\textbf{Training and inference recipes.}
We train the model for 35 epochs with an initial learning rate of $10^{-4}$ and batchsize of $32$. The learning rate is decreased by 10 times at the 20th and 30th epoch respectively. Our approach takes as input $T=16$ frames sampled from each video, and each frame is cropped and resized to a resolution of 224 $\times$ 224. We adopt the multiscale crop strategy with scale factors $[1, 0.875, 0.75]$ when training. During testing, we adopt a center crop in most of our experiments.

\subsection{Ablation Studies}
We first conduct ablation experiments to investigate the effect of different components and design choices in our approach on the two datasets. For Something-Else, we evaluate our approach to the compositional action recognition task. As for Ikea-Assembly, we conduct our evaluation with the data partition provided by the original paper~\cite{ben2021ikea}.

\begin{figure*}[htbp]
    \centering
    \includegraphics[width=0.74\linewidth]{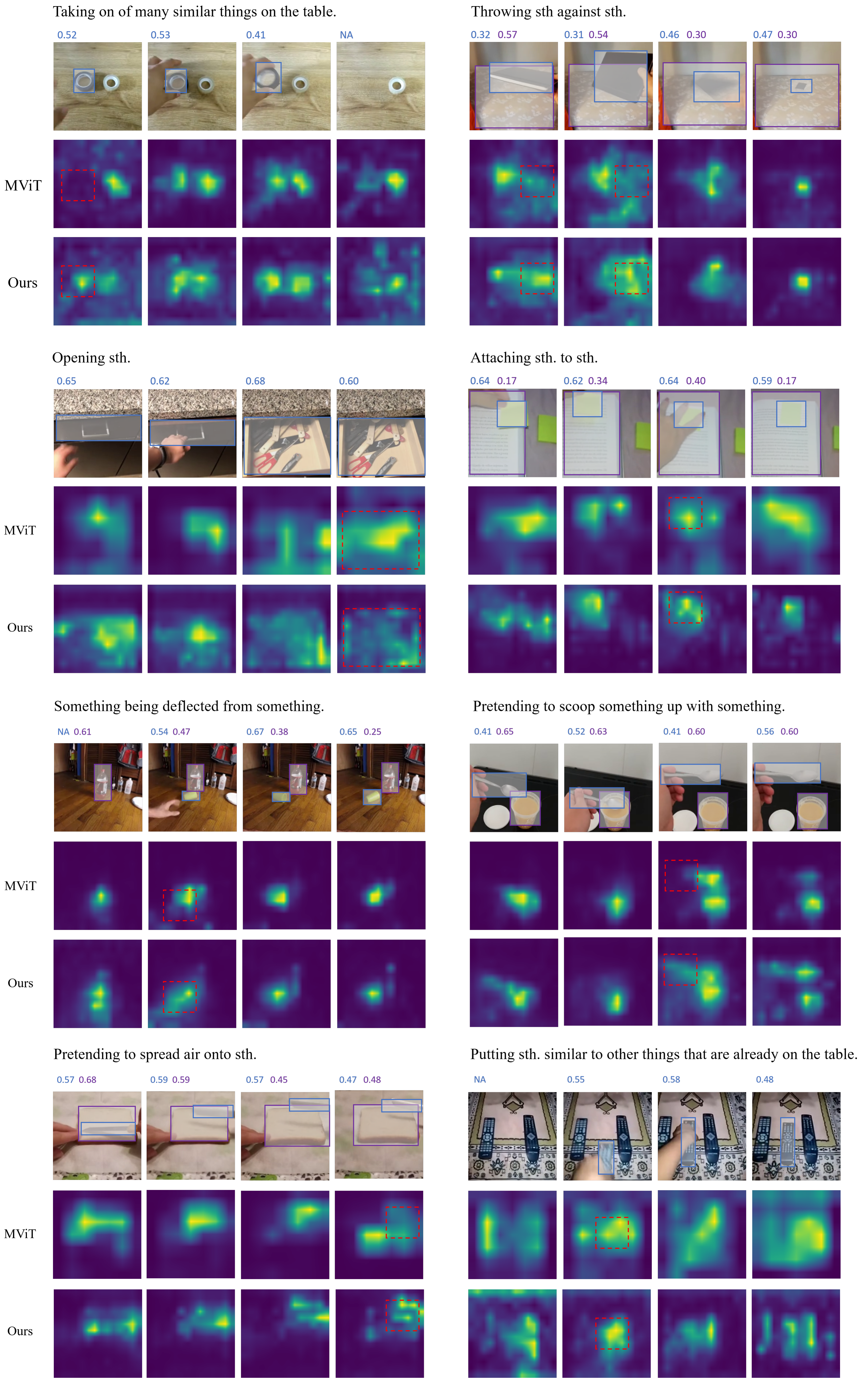}
    \caption{Visualization of the \textbf{Attention Map} comparison and the interactiveness score predicted by \moduleB. We compare the attention region of [CLS] token between \modelname and MViT (the second and third row). We also present the interactiveness score predicted by \moduleB (the first row). The ground truth objects (\ie interactive objects) are drawn by colored boxes, and their corresponding interactiveness scores are displayed above images.}
    \label{fig:visualize}
\end{figure*}

\input{tables/abl_components}

\input{tables/detection_sth}

\input{tables/detection_ikea}

\textbf{Effect of Main Components.}
Table~\ref{tab:components} presents the results of our proposed two main components:~\moduleA and~\moduleB. The baseline model is MViTv2. We first only incorporate~\moduleA with the base model (the second row), which serves as a multitask training that optimizes both object detection and action recognition tasks. We can observe a significant improvement on both datasets, where 2.4\% on Something-Else and 3.3\% macro-recall on Ikea. Moreover, results are further improved by incorporating~\moduleB (the third row), which refines interactiveness scores of candidate objects and injects object-centric information into video representations. These results demonstrate the effectiveness of our proposed modules from two perspectives. Firstly, joint training with object detection is beneficial for action recognition, while it cannot be brought from offline detectors. Secondly, the incorporation of interaction object information is crucial for more accurate interaction modeling.

\textbf{Detection Performance.}
Table~\ref{tab:detection_sth} compares the detection performance between the two commonly used off-the-shelf detectors and our end-to-end approach. For Faster-RCNN~\cite{ren2015faster}, the offline detected boxes are provided by~\cite{materzynska2020something}, which are obtained by fine-tuning a Faster-RCNN detector with ResNet-50~\cite{he2016deep} backbone and Feature Pyramid Network (FPN)~\cite{lin2017feature} on the target dataset. And for DETR~\cite{zhu2020deformable}, which is a video object detection version, it is noticeable that the tokens are extracted with a 3-D backbone MViT to obtain temporal information. We can observe that the detection performance of our~\moduleA performs better than the offline detectors, especially for the generic object category (separately 9.6\% and 3.9\% AP improvements). Since the dataset only annotates interacted objects, these superior results demonstrate the effectiveness of interactive object detection of our method, which benefits from integrating with temporal context information. 
We also present the detection results on the Ikea-Assembly dataset in Table~\ref{tab:detection_ikea}. We notice that our approach performs poorly on small size objects and even fails to make predictions. This can be attributed to the fact that the original video is recorded by high resolution ($1920\times 1080$) and in the third-person perspective, making some objects extremely small and challenging to process when resized to the input scale ($224\times 224$). Scaling up the input resolution of video models to achieve more accurate detection is worth exploring in the future. Nevertheless, jointly training with detections still benefits the base model, as it effectively detects the subject (person) and aggregates the surrounding information to concentrate on the interactive parts.

\textbf{Query Initialization.}
We compare several design choices for initializing queries in \moduleA, and the results are shown in Table~\ref{tab:query}. For the spatial embedding method, ``grid'' represents using the same grid points across frames, while ``random'' represents independent random sampling of points at each frame. We can observe that the ``grid'' initialization only slightly outperforms the baseline, while ``random'' improves the top-1 accuracy by $0.5$ points over it and shows significant improvement in detection metrics. This is due to using the same spatial embedding for queries will encourage them to predict the same location across frames, which prevents them from capturing the large displacements of objects in some cases. Additionally, we observe that incorporating identity embedding further improves the results, with a 1.5\% increase in top-1 accuracy and a 1.0\% increase in top-5 accuracy.

\input{tables/abl_query}

\input{tables/abl_multiscale}

\textbf{Multi-scale features.}
As MViT~\cite{li2022mvitv2} is capable of producing multiscale features in different stages, our approach can easily integrate multi-scale features to enhance detections for small sizes and facilitate gradient propagation in shallow features. The results are presented in Table~\ref{tab:multiscale}. The spatial resolutions of the $9$-th and $2$-nd layer output features are $14\times 14$ and $28\times 28$, respectively. Following~\cite{li2022mvitv2}, we upsample the low-resolution features and align their dimensions with $1\times 1$ convolution. As the size of visual features increases, the overall computation cost also increases accordingly. We find the multiscale features do not provide benefits on Something-Else, but show obvious improvements on Ikea. This could be due to the fact that the Ikea dataset contains more small objects, which benefit more from higher resolution features.

\textbf{Ablation Studies on Different Layers}\label{sec:abl}
We integrate DAIR at different layers of MViTv2~\cite{li2022mvitv2} and compare the performance, computational cost, and parameter count. The results are presented in Table~\ref{tab:layer}. We find that as DAIR is integrated closer to the top layers of the network, its performance decreases more. Specifically, when it is applied to the $15$th layer (the last layer), the entire model degrades into multitask learning, where the object information is not fused into the representation of the video, resulting in the lowest performance. As the feature dimensions of the later layers increase, the parameter count of the model also significantly increases. Additionally, adding DAIR only results in an additional 8\% of computational cost (GFLOPs).
\input{tables/abl_plug_in_layer}

\input{tables/sota_sth}

\begin{figure*}[htbp]
    \centering
    \includegraphics[width=0.95\linewidth]{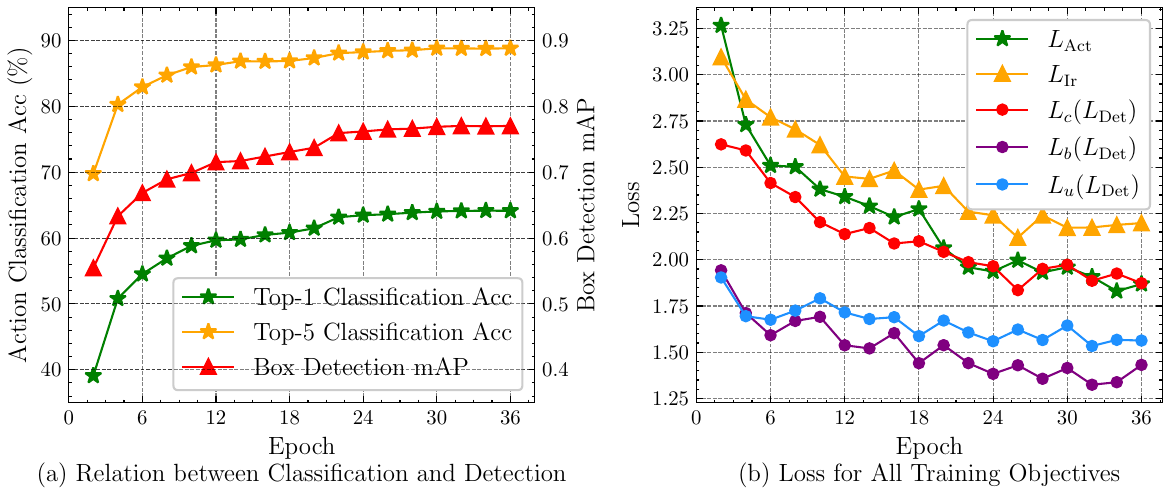}
    \caption{(a) Demonstration of the action classification accuracy (Top-1 Classification Acc and Top-5 Classification Acc) and detection mean average precision (Box Detection mAP) in the validation process, where the x-axis denotes the training epoch. The classification accuracy and detection mAP are in the same trend. (b) The visualization of each training objective in the training process. We visualize the training curve of  and $\mathcal{L}_{Ir}$ and $\mathcal{L}_{Act}$, as well as $\mathcal{L}_{c}$, $\mathcal{L}_{b}$, and $\mathcal{L}_{u}$ in $\mathcal{L}_{Det}$. It demonstrates that all these training objectives could optimize classification and detection simultaneously.}
    \label{fig:curve}
\end{figure*}
\vspace{-3pt}

\subsection{Comparison with State-of-the-arts} 
We compare our~\modelname with the state-of-the-art on both Something-Else and Ikea-Assembly.

\noindent\textbf{Something-Else. } 
Table~\ref{tab:sota-sthelse} reports the results of the compositional and few-shot task. The second block in Table~\ref{tab:sota-sthelse} compares different object-centric approaches with detected bounding boxes. We can observe that our method achieves the state-of-the-art results, and outperforms the previous best method with 1.0\% top-1 accuracy and 1.3\% top-5 accuracy on the challenging compositional task.
To elaborate further, our main competitors can be categorized into two groups: the first is multi-branch models that process RGB inputs and box inputs separately (\eg STIN~\cite{materzynska2020something} and STLT~\cite{radevski2021revisiting}), and for a fair comparison, we reproduce their results by using the same RGB model as ours (\ie MViT). The second group consists of transformer-based models, such as ORViT~\cite{herzig2022object} and SViT~\cite{ben2022bringing}, where the former jointly encodes box inputs and patch tokens, while the latter is a multitask model that trains object tokens responsible for detection on image data with object annotations and aligns them with video object tokens. Based on our superior results, we can draw the following conclusions: \textrm{i}) in terms of incorporating object-centric information into the video representation, our joint training approach is more effective than the separate late fusion strategy adopted by multi-branch methods. \textrm{ii}) compared to existing transformer-based methods, our approach provides a more unified object-centric recognition pipeline with better performance under a realistic setting.

\noindent\textit{More Analysis under the Oracle setting. } Most previous works evaluate their object-centric models on the oracle setting (accessing manually annotated boxes in both training and testing), in order to fully demonstrate the object-level reasoning capabilities. We also consider comparing our method under this oracle setting. We replace the interactive object refining step in~\moduleB with ground truth boxes to filter the object candidates predicted by~\moduleA. Specifically, we adopt two schemes: ``GT box'' is to perform bipartite matching between ground truth boxes and our predicted boxes, and ``GT num'' refers to the selection of predicted boxes based on their confidence scores that equal the number of ground truth boxes at each frame. The results are presented in the third block of Table~\ref{tab:sota-sthelse}. Surprisingly, by incorporating the prior of interactive objects, our method achieves significant improvements in both the ``GT box'' and ``GT num'' schemes and surpasses the state-of-the-art. It is worth emphasizing that we only use ground truth boxes for filtering object proposals and do not employ specialized design for the subsequent object reasoning. These results strongly support the effectiveness of interactive object knowledge in object-centric action recognition, which is overlooked in previous works. 

\input{tables/sota_ikea}

\noindent\textbf{Ikea-Assembly. } Table~\ref{tab:sota-ikea} also shows that we surpass the state-of-the-art on Ikea-Assembly. For a fair comparison, we re-implement some object-centric models (\eg STIN~\cite{materzynska2020something}, STRG~\cite{wang2018videos}) using the same backbone, which is served as strong baselines. Notably, our model surpasses the previous best method MGAF~\cite{kim2021motion} by 4.1 points on macro-f1 recall and 9.6 points on micro-f1 accuracy. It is worth noting that our model only utilizes the ``pseudo ground truth'' of object annotations released by~\cite{ben2021ikea} for training. This indicates that our approach is robust to some noisy supervision cases and can be extended to weaker annotations potentially, such as manually annotating a few key frames only.

\subsection{Visualization}
Figure~\ref{fig:visualize} visualizes the attention map of~\modelname and MViT, and the interactiveness score predicted by~\moduleB. From the attention maps, it can be seen that~\modelname attends more to the regions where the interaction exists. For instance, in the first example,~\modelname entirely focuses on the left object that is being taken away. In the second example, there is more attention on the area where the hand and the ``book'' come into contact. As for the interactiveness score calibrating, it can be observed from the second example that the score of the book increases as its movement progresses. In contrast, the score of the background bed relatively decreases, even though it occupies a much larger space compared to the ``book''.
As mentioned above, after integrating \modelname, the [CLS] token pays more attention to the interacted objects highlighted by the red dashed boxes.
Furthermore, we observe that the ground truth object boxes (in blue and purple) usually have higher interactiveness scores. Objects that are passively involved in the action have their interaction score increase gradually throughout the action, as demonstrated by the example of the \textit{book} in the \textit{attaching sth. to sth.} action.

\subsection{Demonstration of Classification and Detection}\label{sec:clsdet}
The depicted Figure~\ref{fig:curve} succinctly illustrates the concurrent evolution of action classification and object detection throughout the training process. Notably, the close alignment of top-1 and top-5 classification accuracy with detection mean average precision underscores a cooperative optimization between these tasks. The parallel performance trends throughout validation epochs signify a shared learning trajectory, emphasizing the interdependence of action classification and object detection. This harmonious relationship enhances the model's overall capability, showcasing that improvements in one task contribute synergistically to advancements in the other. In essence, the cooperative optimization of the two tasks proves essential in achieving a robust and comprehensive solution for egocentric action recognition.

In (b) we delve deeper into the training process by visualizing each specific training objective. We portray the training curves for $\mathcal{L}{Ir}$, $\mathcal{L}{Act}$, as well as $\mathcal{L}{c}$, $\mathcal{L}{b}$, and $\mathcal{L}{u}$ within the broader context of $\mathcal{L}{Det}$. This comprehensive visualization illuminates the training dynamics associated with individual objectives and underscores their simultaneous impact on both classification and detection tasks. The collective optimization of these diverse training objectives further emphasizes the model's capacity to refine its understanding across multiple dimensions, ultimately contributing to the cohesive enhancement of both classification and detection performance throughout the training iterations.

%% file: tables/abl_components.tex
\begin{table}[]
\centering
\caption{Effect of main components of our approach on both Something-Else and Ikea-Assembly datasets.~\moduleA: Patch-based object Decoding.~\moduleB: Interactive objects Refining and Aggregation.}
\scalebox{1.25}{
\begin{tabular}{@{}lcccc@{}}
\toprule
 & \multicolumn{2}{l}{SomethingElse} & \multicolumn{2}{l}{Ikea-Assmbly} \\ 
 & Top-1 & Top-5 & Macro & Micro \\ \midrule
MViTv2 & 63.4 & 87.8 & 47.6 & 79.2 \\
+~\moduleA & 65.8 & 89.2 & 50.9 & 81.5 \\
\quad +~\moduleB & \textbf{66.3} & \textbf{89.6} & \textbf{53.2} & \textbf{82.0} \\ \bottomrule
\end{tabular}
}
\label{tab:components}
\end{table}

%% file: tables/detection_sth.tex
\begin{table}
\centering
\caption{Detection performance (AP@$50$ in \%) comparison on Something-Else. ``Sub'' and ``Obj'' are short for ``Subject'' and ``Object'' respectively.}
\scalebox{1.25}{
\begin{tabular}{@{}lccc@{}}
\toprule
Method & Sub & Obj & mAP \\ \midrule
Faster-RCNN~\cite{ren2015faster} & 80.7 & 65.6 & 73.1  \\
DETR~\cite{zhu2020deformable} & 80.5 & 71.3 & 75.9 \\
DAIR (Ours) & 80.7 & 75.2 & 78.0  \\ \bottomrule
\end{tabular}
}
\label{tab:detection_sth}
\end{table}

%% file: tables/detection_ikea.tex
\begin{table}
\centering
\caption{Detection performance (AP@$50$ in \%) on Ikea-Assembly. ``t'' and ``p'' is short for ``top'' and ``panel'' respectively.}
\scalebox{1.15}{
\begin{tabular}{@{}llllllllc@{}}
\toprule
\rotatebox{90}{person} & \rotatebox{90}{table-t}  & \rotatebox{90}{leg} & \rotatebox{90}{shelf} & \rotatebox{90}{side-p} & \rotatebox{90}{front-p} & \rotatebox{90}{bottom-p} & \rotatebox{90}{rear-p} & mAP \\ \midrule
99.1 & 57.6 & 2.3 & 22.9 & 0.4 & 8.0 & 22.5 & 1.7 & 26.8  \\ \bottomrule
\end{tabular}
}
\label{tab:detection_ikea}
\end{table}

%% file: tables/abl_query.tex
\begin{table}[]
\centering
\caption{Results on Something-Else of different query initialization methods. Default choices of our module are colored in \colorbox[HTML]{EFEFEF}{gray}.}
\scalebox{1.18}{
\begin{tabular}{@{}cccccc@{}}
\toprule
\multicolumn{2}{c}{Spatial Emb.} & \multirow{2}{*}{Identity Emb.} & \multirow{2}{*}{Top-1} & \multirow{2}{*}{Top-5} & \multirow{2}{*}{mAP} \\ \cmidrule{1-2}
\multicolumn{1}{c}{Grid} & \multicolumn{1}{c}{Random} &  &  &  &  \\ \midrule
 \cmark & \xmark & \xmark & 63.8 & 88.1 & 48.8 \\
 \xmark & \cmark & \xmark & 64.3 & 88.2 & 70.7 \\
 \rowcolor[HTML]{EFEFEF}
 \xmark & \cmark & \cmark  & \textbf{65.8}  & \textbf{89.2} & \textbf{78.0} \\ \bottomrule
\end{tabular}
}
\label{tab:query}
\end{table}

%% file: tables/abl_multiscale.tex
\begin{table}[]
\centering
\caption{Results of the utilization of multi-scale features when decoding objects.}
\scalebox{1.18}{
\begin{tabular}{@{}ccccccc@{}}
\toprule
Layers & \multicolumn{2}{c}{Something-Else} & \multicolumn{2}{c}{Ikea-Assmbly} & \multirow{2}{*}{GFLOPs} \\ 
 & Top-1 & Top-5 & Macro & Micro &  &  \\ \midrule
baseline & 63.4 & 87.8 & 47.6 & 79.2 & 64.5  \\
9 & \textbf{66.3} & \textbf{89.6} & 51.4 & 80.9 &  69.9 \\ 
2,9 & 66.0 & 89.4 & \textbf{53.2} & \textbf{82.0} & 78.4  \\ \bottomrule
\end{tabular}
}
\label{tab:multiscale}
\end{table}
\vspace{-3pt}

%% file: tables/abl_plug_in_layer.tex
\begin{table}[t]
\centering
\caption{Applying DAIR into the different layer.}
\scalebox{1.25}{
\begin{tabular}{@{}cllcc@{}}
\toprule
\#Layer & \multicolumn{1}{c}{Top-1} & \multicolumn{1}{c}{Top-5} & \multicolumn{1}{c}{GFLOPs} & \multicolumn{1}{c}{Params} \\ \midrule
baseline & 63.4 & 87.8 & 64.5 & 36.4 \\
\rowcolor[HTML]{EFEFEF}
9 & \textbf{66.3} & \textbf{89.6} & 69.9 & 46.6 \\
13 & 65.7 & 89.5 & 69.8 & 46.6 \\
15 & 64.4 & 88.8 & 67.7 & 60.9 \\ \bottomrule
\end{tabular}
}
\label{tab:layer}
\end{table}

%% file: tables/sota_sth.tex

\begin{table}
\centering
	\caption{Compositional and few-shot action recognition on the ``Something-Else'' dataset.}
	\label{tab:sota-sthelse}
	\scalebox{1.07}{
		\begin{tabular}{lcccc}
			\toprule
			\multicolumn{1}{c}{\multirow{2}{*}{Method}} & \multicolumn{2}{l}{Compositional} &  \multicolumn{2}{c}{Few-Shot\footnotemark[1]} \\ \cmidrule(l){2-5} 
			\multicolumn{1}{c}{} & Top-1 & Top-5 & 5-Shot & 10-Shot \\ \cmidrule(r){1-5}
			MFormer~\cite{patrick2021keeping} & 60.2 & 85.8  & 28.9 & 33.8   \\
			MViT~\cite{li2022mvitv2} & 63.3 & 87.5 & 32.7 & 40.2   \\ \cmidrule(r){1-5} 
			\multicolumn{5}{l}{\cellcolor[HTML]{EFEFEF}\textbf{detection}} \\
			STIN~\cite{materzynska2020something} (w/~MViT~\cite{li2022mvitv2})  & 48.2 & 72.6 & 23.7 & 27.0 \\
			STLT~\cite{radevski2021revisiting} (w/~MViT~\cite{li2022mvitv2})  & 56.9 & 82.5 & 27.1 & 33.9 \\ 
            MGAF~\cite{kim2021motion} & 61.2 & 83.3 & - & - \\
			SViT~\cite{ben2022bringing}  & 65.8 & 88.3 & 34.4 & \textbf{42.6} \\
			Ours & \textbf{66.8} & \textbf{89.6} & \textbf{34.8} & \textbf{42.6}  \\ 
			\multicolumn{5}{l}{\cellcolor[HTML]{EFEFEF}\textbf{oracle}} \\
			MGAF~\cite{kim2021motion} & 68.0 & 88.7 & - & - \\
			ORViT~\cite{herzig2022object} & 69.7 & 91.0 & 33.3 & 40.2 \\
                Ours + GT Num & 69.7 & 91.3 & 36.3 & 42.8 \\
                Ours + GT Match & \textbf{72.3 }& \textbf{93.0} & \textbf{38.8} & \textbf{44.7} \\
			\bottomrule
		\end{tabular}
	}
\end{table}
\footnotetext[1]{The symbol ``-'' in Few-Shot column indicates that the results are not reported in the original paper and the code is not publicly available for reproduction.}

%% file: tables/sota_ikea.tex

\begin{table}
\centering
	\caption{Results on the original task of the Ikea-Assembly dataset.}
	\scalebox{1.25}{
        \begin{tabular}{lcc}
			\toprule
			Method & Macro & Micro \\ \midrule
			MViT~\cite{li2022mvitv2} & 47.6 & 79.2 \\
                STIN~\cite{materzynska2020something} (w/~MViT)  & 48.0 & 81.0   \\
			MGAF\cite{kim2021motion}  & 49.1 & 72.4  \\ 
			STRG~\cite{wang2018videos} (w/~MViT) & 52.3 & 79.8  \\
			Ours &  \textbf{53.2} & \textbf{82.0}  \\ \bottomrule
	   \end{tabular}
    }
	\label{tab:sota-ikea}
\end{table}

%% file: sections/5_conclusion.tex
\section{Conclusion}

In this work, we propose~\modelname for performing object-centric action recognition in an end-to-end manner. We devise three consecutive modules that extract more accurate interactive object information from both the spatio-temporal representation at patch-level and the contextual relationships at instance-level. In addition, we fuse the interactive object information with video patch tokens through joint self-attention, thereby enhancing the video representation. Our extensive experiments demonstrate the importance of incorporating the interactiveness before and the effectiveness of the proposed~\modelname, which achieves state-of-the-art performance on both Something-Else and Ikea-Assembly challenging benchmarks.